\documentclass[onecolumn,technote,twoside]{IEEEtran}
\usepackage[utf8]{inputenc}
\usepackage{t1enc,amsmath,amsfonts,epsfig,graphicx,color}
\usepackage[active]{srcltx} 

\def\BibTeX{{\rm B\kern-.05em{\sc i\kern-.025em b}\kern-.08em
    T\kern-.1667em\lower.7ex\hbox{E}\kern-.125emX}}
\usepackage[active]{srcltx} 

\usepackage{bm}
\usepackage{cite}
\usepackage{amsmath,amssymb,amsfonts}
\usepackage{algorithmic}
\usepackage{graphicx}
\usepackage{textcomp}

\usepackage{empheq}
\usepackage{booktabs}

\renewcommand{\baselinestretch}{1 } 

\newcommand{\bi}[1]{\textbf{\textit{#1}}}

\input{def11.set}

\title{Mirror Descent Using the Tempesta Generalized Multi-parametric Logarithms}

\author{ Andrzej CICHOCKI \\
Systems Research Institute of Polish Academy of Science, Newelska 6, 01-447 Warsaw, Poland,\\
 and  Nicolaus Copernicus University in Torun (UMK),  Poland,\\
visiting Professor in Global Innovation Research Institute of Tokyo University of Agriculture and Technology, 2-24-16 Naka-cho, Koganei-shi, Tokyo, Japan,\\
Riken AIP, 103-0027 Tokyo, Nihonbashi, 1 chome-4-1, Japan, (e-mail: a.cichocki@riken.jp)}

\begin{document}

\maketitle


\begin{abstract}

In this paper, we develop a wide class  Mirror Descent (MD) algorithms,  which play key role in machine learning. For this purpose we formulated the constrained optimization problem, in which we exploits  the Bregman divergence with the Tempesta multi-parametric  deformation logarithm as a link function. This link function  called also mirror function defines the mapping between the primal and dual spaces and is associated with a very-wide (in fact, theoretically infinite) class of generalized  trace-form entropies. In order to derive novel MD updates, we estimate generalized  exponential function, which closely approximates the inverse of the multi-parametric Tempesta generalized logarithm.  The shape  and properties of the Tempesta logarithm and its inverse-deformed exponential functions can be tuned by several hyperparameters. By learning these hyperparameters,  we can adapt to distribution or geometry of training  data, and we can adjust them to achieve desired properties of MD algorithms. The concept of applying  multi-parametric logarithms  allow us to generate a new wide and flexible family of  MD and  mirror-less MD updates.
\end{abstract}


\section{Introduction}

Mirror descent (MD) is a powerful optimization approach used in deep learning, machine learning and optimization problems, especially in online learning, that extends the gradient descent by allowing for non-Euclidean geometries and flexible distance metrics through the use of a 'mirror map' or a potential function, enabling more effective optimization in various scenarios \cite{Nemirowsky,Beck2003,MD1,MD2,EGSD,han2022,han2025,shalev2011}.  One reason for their success is the fact that mirror descent can be adapted to fit the geometry of the optimization problem at hand by choosing suitable strictly convex potential functions (mirror maps). In essence, mirror descent offers a powerful framework for optimization by leveraging dual spaces and Bregman divergence, allowing for more tailored and potentially faster convergence in various settings,
including convex optimization problems and also in non-convex settings \cite{Beck2003,MD1,amid2022,wang2024}.

Mirror descent typically employs a Bregman divergence that serves as a proximal (regularization) function of an optimized loss function. The advantage of using mirror descent over standard gradient descent methods is that it takes into account the geometry of the problem through a link function $f(x)$, which has to be strictly concave. By using different Bregman divergences, mirror descent can be tailored to the specific geometry of the optimization problem, potentially leading to faster convergence or better solutions. If the distance-generating function is $F(w) =  =(1/2)||\bw||_2^2$, then the induced Bregman distance is the squared Euclidean distance, and mirror descent recovers the standard gradient descent. Other Bregman distances will  be used to optimize over different geometries, especially the Tempesta multi-parametric logarithm \cite{tempesta2015,tempesta2016}.

In online learning, mirror descent can be used for updating weights in a multiplicative fashion, resulting in exponential gradient (EG) descent \cite{EG,KW1995,MD2,Cichocki_Cruces_Amari,Cichocki2024,Cichocki2025}. There are also variations of mirror descent that is used in stochastic optimization and includes stochastic gradient descent (SGD) as a special case \cite{azizan2021,EGSD}.
,
The main objective of  this paper is development generalized and flexible MD updates, which arise under a broader principle: by exploiting  multi-parametric link functions \cite{tempesta2015,tempesta2016}, which allows  us to adapt to  any data with various  geometry and probability distributions.
The  developed MD updates can be adapted to the geometry of the optimization problem, potentially leading to faster convergence and better performance, especially in high-dimensional spaces.
The MD can handle non-smooth objectives and non-Euclidean norm constraints, which are limitations of standard gradient descent (GD).


 \subsection{Preliminaries: Mirror Descent and Standard Exponentiated Gradient (EG) Update}

 {\bf Notations}. Vectors are denoted by boldface lowercase letters, e.g., $\bw \in \Real^N$, where for any vector $\bw$, we denote its $i$-th entry by $w_i$. For any vectors $\bw, \bv \in \Real^N$.  The $N$-dimensional real vector space with nonnegative real numbers is denoted by $\Real^N_+$. We define the Hadamard product as $\bw \odot \bv = [w_1 v_1, \ldots, w_N v_N]^T$ and $\bw^{\alpha} = [w_1^{\alpha}, \ldots, w_N^{\alpha}]^T$. All operation for vectors like multiplications and additions are performed componentwise. The function of a vector is also applied for any entry of the vectors, e.g., $f(\bw) = [f(w_1),f(w_2),\ldots, f(w_N)]^T$.   We let $\bw (t)$ denote the weight or parameter vector as a function of time $t$. The learning process advances in iterative steps, where during step $t$ we start with the weight vector $\bw (t) = \bw_t$ and update it to a new vector $\bw(t+1) = \bw_{t+1}$. We define $[x]_+ = \max\{0,x\}$, and the gradient of a differentiable cost function as $ \nabla_{\bi w} L(\bw) = \partial L(\bw)/\partial \bw = [\partial L(\bw)/\partial w_1, \ldots, \partial L(\bw)/\partial w_N]^T $. In contrast to generalized logarithms defined later the classical  natural logarithm will be denoted by $\ln (x)$.\\

{\bf Problem Statement:}
Suppose, we want to minimize a loss/cost function $L(\bw)$,  with respect the weight vector $\bw=[w_{1},\ldots,w_{N}]^T\in \mathbf{R}_+^N$, i.e., we want to solve the following
constrained optimization problem:
\be
	\bw_{t+1} = \argmin_{{\bi w} \in \mathbf{R}_+^N} \left\{ L(\bw )+ \frac{1}{\eta} D_F(\bw || \bw_t)  \right\}, \;\; \text{s.t.} \;\; \bw_i>0,\; \forall i \;\; ||\bw||_1=\sum_{i=1}^{N}, w_{i}=1,
	\label{optim1}
\ee
where  $L(\bw)$ is a differentiable loss function,  $\eta > 0$ is the learning rate and
$D_F(\bw || \bw_t)$ is the Bregman divergence \cite{Bregman1967}.

The Bregman divergence is  defined as \cite{Bregman1967,MD1,Amari2009}
\be
D_F(\bw || \bw_t) = F(\bw) - F(\bw_t) -  (\bw-\bw_t)^T f(\bw_t),
\label{Bregman1}
\ee
where  generative function (mirror map) $F(\bw)$ is a continuously-differentiable, strictly convex function  defined on the convex domain, while $f(\bw)= \nabla_{\bi w} F(\bw)$, is strictly concave  function, which is called the link function.

 The optimization problem (\ref{optim1}) can be solved iteratively  by applying 
 so-called prox or implicit MD update \cite{MD1,shalev2011,Cichocki2025}
 \begin{empheq}[box=\fbox]{align}
\bw_{t+1} &= f^{(-1)} \left[ f(\bw_t) - \eta \nabla_{\bi w} L(\bw_t)\right] 
\label{f-1fDT}\\
\bw_{t+1} &\leftarrow \bw_{t+1} /||\bw_{t+1}||_1 \nonumber
\end{empheq}
or optionally  by applying MirrorLess MD  (MMD) update \cite{gunasekar2021}, normalized to unit-length $\ell_1$ norm:
\begin{empheq}[box=\fbox]{align}
\bw_{t+1} =& \left[\bw_t  -\eta \diag \left\{\left(
\frac{d\,f(\bw_t)}{d \bw_t}\right)^{-1}\right\} \nabla_{\bi w} L(\bw_t)\right]_+ 
 \label{diagMD} \\
\bw_{t+1} &\leftarrow \bw_{t+1} /||\bw_{t+1}||_1 \nonumber
\end{empheq}
where $ \displaystyle \diag \left\{ \left(\frac{d\,f(\bw)}{d \bw}\right)^{-1}\right\} = \diag \left\{ \left(\frac{d\,f(\bw)}{d w_1} \right)^{-1}, \ldots, \left(\frac{d\,f(\bw)}{d w_N}\right)^{-1} \right\} $.

Mirror descent (\ref{f-1fDT}) operates by explicitly distinguishing between the primal (original) and dual spaces of the optimization problem. It establishes a bijection (one-to-one correspondence) between these two spaces using a link function (in our case  family of generalized logarithms).
This allows the MD updates to work in a dual space, effectively performing gradient descent on the dual problem, which can offer advantages in many scenarios.

On the other hand, MMD (\ref{diagMD}) can be considered as a specific way of u deriving Mirror Descent, an optimization algorithm, without explicitly relying on a dual space or a mirror map. It focuses on a primal-only perspective, viewing Mirror Descent as a "partial" discretization of gradient flow on a Riemannian manifold. 

%
The key step in our  approach is  a suitable  choice of generalized multi-parametric logarithm  as a flexible, parameterized link functions $f(\bw)$,  which allow us to adopt to various distributions  of training  data.
Although, some extensions of MDD/MD updates has been already proposed included one-parametric Tsallis or Kaniadakis  logarithm, and two-parameters Euler logarithm \cite{Cichocki_Tanaka_Cruces2025,Cichocki2025},  however, to our best knowledge  the Tempesta generalized multiparametric logarithm has been not investigated neither applied  till now for MD/MDD  updates \cite{tempesta2015,tempesta2016}.

\section{The Tempesta Functional Multiparametric Logarithm and its Basic Properties}

In this paper we investigate the application of the  generalized Tempesta logarithm \cite{tempesta2015} as a link function $f(x)$ in the mirror-less MD (MMD)  and mirror descent (MD):
\be
\displaystyle \log^{Te}_{\phi,\alpha,\sigma}(x)
=\left\{
	\begin{array}{cl}
\displaystyle \frac{1}{1 -\alpha \phi'(\alpha} \left[\frac{\phi(\alpha x^{-\sigma}) - x^{-\sigma}}{\sigma}+ \frac{1- \phi(\alpha)}{\sigma} \right] & \text{for} \; x>0, \;\; \sigma \neq 0, \; \;  \alpha \phi'(\alpha) \neq 1,\\
 \\
 \displaystyle \frac{x^{1-q}-1}{1-q} & \text{for} \;\;x>0, \;\; \alpha=0, \; \sigma=q-1,\\
 \\
		\ln (x) & \text{for} \;\; x>0 \;\; \text{and} \;\; \sigma=0,
	\end{array}
	\right.
\label{logTe}
 \ee
where function $\phi(x)$ is any differentiable  function,  which in general satisfies the following condition
\be
\frac{1 -\alpha \phi'(\alpha x^{\sigma})- \alpha^2 x^{\sigma} (\sigma/(1+\sigma)) \phi''(\alpha x^{\sigma})}{1 -\alpha \phi'(\alpha)} >0.
\ee
So defined logarithm function is associated  to very wide class of entropies \cite{tempesta2015}:
\be
S^{Te}(\bp) = k_B \sum_{i=1}^{W} p_i \log^{Te}_{\phi,\alpha,\sigma} (1/p_i)
\ee
The Tempesta logarithm can be represented approximately by the following
power series \cite{tempesta2015,tempesta2016}:
\be
\log_{\phi,\alpha,\sigma}^{Te}(x) &=& \ln(x) - \sigma \frac{\alpha^2 \phi''(\alpha) + \alpha  \phi'(\alpha) -1}{2(-1 +\alpha \phi'(\alpha))}\ln^2 (x) + \sigma^2 \frac{\alpha^3 \phi'''(\alpha) + 3\alpha^2  \phi''(\alpha) + \phi'(\alpha) -1}{6(-1 +\alpha \phi'(\alpha))} \ln^3{x} + \cdots \\
&=& \ln(x) + \frac{1}{2} a_1 \ln^2(x) + \frac{1}{6} a_2 \ln^3(x) + \cdots
\label{logTS}
\ee
For suitable chosen range of hyper-parameters $\alpha \in [\alpha_{\min}, \alpha_{\max} ]$, $\sigma \in [\sigma_{\min}, \sigma_{\max} ]$  and  generating function $\phi(\cdot)$  the Tempesta logarithm is strictly concave function  and satisfies the following fundamental properties:

  \begin{itemize}

 \item

 Domain $\log_{\phi,\alpha,\sigma}^{Te}(x) $:  $\Real^+ \rightarrow \Real$\\

 \item

 Monotonicity: $\displaystyle \frac{d \log_{\phi,\alpha,\sigma}^{Te}(x)}{dx} >0$ \\

 \item

 Concavity:  $\displaystyle \frac{d^2 \log_{\phi,\alpha,\sigma}^{Te}(x)}{dx^2} < 0$ \\

 \item

 Scaling and Normalization: $\log_{\phi,\alpha,\sigma}^{Te}(1)=0$, \;\;  $\displaystyle \frac{d \log_{\phi,\alpha,\sigma}^{Te}(x)}{dx}\vert_{x=1} =1$\\

 \item

Remark:  Self-Duality property  is in general not satisfied,
 except some special cases, i.e., $\log_{\phi,\alpha,\sigma}^{Te}(1/x) \neq - \log_{\phi,\alpha,\sigma}^{Te}(x)$

 \end{itemize}

 It important to emphasise that the Tempesta logarithm  is a magic formula,  which is very flexible and general, in the sense that allow us to construct as special cases almost all known and important generalized logarithms and theoretically infinitely many new ones. We illustrate this by only a few examples.
,
\subsection{Generalized one parameter and bi-parametric logarithms}

By selecting $\phi(x) = ax -c $ for arbitrary real parameters $a, c \in \Real$,
the generalized logarithm (\ref{logTe}) simplifies to the Tsallis $q$-logarithm for $\sigma =q-1$  and  any $\alpha \neq 1/a$
 \be
	\log^T_{q}(x)=\left\{
	\begin{array}{cl}
		\displaystyle \frac{x^{\,1-q}-1}{1-q}  & \text{if} \;\;  x>0 \;\; \text{and} \;\;  q \neq 1,\\
 \\
		\ln (x) & \text{if} \;\; x>0 \;\; \text{and} \;\; q=1.
	\end{array}
	\right.
\label{logT}
\ee
which inverse function take the form
\be
 \label{defexpq}
	\exp^T_{q}(x)=\left\{
	\begin{array}{cl}
		\displaystyle [1+ (1-q) x]_+^{1/(1-q)} & q \neq 1\\
 \\
		\exp(x) & q=1
	\end{array}
	\right.
\ee
For $\phi(x) =1/x$, we obtain the extended Kaniadakis logarithm for $\sigma=\kappa \in [-1,1]$
\be
\log_{1/x} = \frac{1}{(1+\alpha) \sigma} \left[ x^{\sigma} - \alpha x^{-\sigma} + \alpha -1\right],
\;\; \text{if} \;\; x>0 \;\; \alpha \neq -1, \;\; \sigma  \neq 0,
\ee
which reduces to the Kaniadakis $\kappa$-logarithm for $\alpha=1$ and $\sigma=\kappa$
\be
 \label{deflogk}
	\log^K_{\kappa}(x)=\left\{
	\begin{array}{cl}
		\displaystyle \frac{x^{\kappa}- x^{-\kappa}}{2 \kappa} = \frac{1}{\kappa} \sinh (\kappa \ln (x)) & \text{if} \;\;  x>0 \;\;
\text{and} \;\;  0  <\kappa^2 <1\\
\\
		\ln (x) & \text{if} \;\; x>0 \;\; \text{and} \;\; \kappa=0.
	\end{array}
	\right.
\ee
The inverse function of the Kaniadakis $\kappa$-logarithm is the deformed exponential function $\exp^K_{\kappa} (x)$, defined as follows:
\be
 \label{defexpk}
	\exp^K_{\kappa}(x)= \exp \left( \int_{0}^{x}\frac{d y}{\sqrt{1+\kappa^2 y^2}}\right) =\left\{
	\begin{array}{cl}
		\displaystyle \left(\sqrt{1+\kappa^2 x^2} + \kappa x\right)^{1/\kappa}
= \exp\left(\frac{1}{\kappa} \; \text{arsinh} (\kappa x) \right)&  0 < \kappa^2 \leq 1 \\
\\
		\exp(x) & \kappa =0
	\end{array}
	\right.
\ee

\subsection{Generalized three-parameters Logarithms}

Let consider the following generating function:
\be
\phi(x) &=& x^r \left[ (\lambda x)^{\kappa} - (\lambda x)^{-\kappa} \right] +x \nonumber \\
&=& \lambda^{\kappa} x^{r+\kappa} -\lambda^{-\kappa} x^{r-\kappa} +x
\ee
In the special case  for $\alpha=1$ and $\sigma=-1$,we obtain the three parameters  generalized logarithm investigated by Kaniadakis  \cite{kaniadakis2009}
\be
\displaystyle \log_{\kappa,r,\lambda} (x) = \frac{\lambda^{\kappa} x^{r+\kappa} - \lambda^{-\kappa} x^{r-\kappa} - \lambda^{\kappa} + \lambda^{-\kappa} }{(r+\kappa) \lambda^{\kappa} - (r-\kappa) \lambda^{-\kappa}}, \;; \text{for}\;\; x>0, \;\; \lambda >0, \kappa \in [-1,1],  \;\; -|\kappa| < r < |\kappa|.
\ee
In the special case for $\lambda=1$ the 3-parameter logarithm simplifies to the Kaniadakis-Lassia-Scarfone (KLS) \cite{kaniadakis2004,kaniadakis2005}, which is equivalent to the Sharma-Mittal-Taneja (SMT) logarithm  \cite{sharma1975,mittal1975}
\be
\log_{\kappa,r} (x) = x^r  \; \frac{x^{\kappa}- x^{-\kappa}}{2 \kappa}, \;\; x>0, \;\; r  \in \Real,\; \; \;\;\text{for}  \;\;\
-|\kappa| \leq r \leq \1/2 - |1/2- |\kappa| |.
\ee
Summarizing, the $(\kappa,r)$-logarithm can be described as follows
\be
 \label{deflogKLS}
	\log^{KLS}_{\kappa,r}(x)=\left\{
	\begin{array}{cl}
		\displaystyle \frac{x^{r+\kappa}- x^{r-\kappa}}{2 \kappa} & \text{if} \;\; x>0 \;\;
\text{for} \;\; r  \in \Real,\; \;\text{and}  \;\;\
 -|\kappa| \leq r \leq |\kappa|,\\
\\
\displaystyle \log^K_{\kappa} (x) = \frac{x^{\kappa} - x^{-\kappa}}{2 \kappa} & \text{if} \;\; x>0 \;\;
\text{for} \; r=0, \; \kappa \in [-1,1], \;\; \kappa \neq 0, \\
\\
\displaystyle \log^T_q (x) = \frac{x^{1-q} -1}{1-q}, & \text{if} \;\; x>0 \;\;
\text{for} \; r= \kappa=(1-q)/2, \;\; q \neq 1, \; q>0,\\
\\
		\ln (x) & \text{if} \;\; x>0 \;\; \text{and} \;\; r=\kappa=0.
	\end{array}
	\right.
\ee

Another interesting generating  function  takes the following form:
\be
\displaystyle \phi(x) = \frac{x^a - x^b}{a x^b -b x^a}+x,
\ee
which allows us to construct the following three parameters logarithm for $\alpha=1$
\be
 \log_{\phi,\sigma,a,b}(x) = - \frac{1}{\sigma} \left[ \frac{x^{-a \sigma} - x^{-b \sigma}}{a x^{-b \sigma} - b x^{-a \sigma} }\right]
 \ee
 For $\sigma=-1$ the above logarithm   simplifies to Hanel-Thurner-Gell-Mann (HTG) two parameter logarithm \cite{hanel2012}
 \be
\label{HTG1}
	\log_{a,b}^{HTG} (x)=\left\{
	\begin{array}{cl}
		\displaystyle \frac{x^{a-b} -1}{a-b \; x^{a-b}}, & x>0, \;\; a\neq b\\
  \\
	\displaystyle	\frac{\ln(x)}{1-a\ln(x)} &  x>0, \;\; a= b,
	\end{array}
	\right.
\ee
which inverse can be explicitly expressed as
\be
 \label{defexpHT}
	\exp^{HTG}_{a,b}(x) =\left\{
	\begin{array}{cl}
		\displaystyle  \left(\frac{1+a\;x}{1+b\;x} \right)^{1/(a-b)} & a \neq b\\
\\
	\displaystyle	\exp\left(\frac{x}{1+ax}\right), & a=b
	\end{array}
	\right.
\ee
%
The HTG-logarithm has the following basic properties:
\be
\log_{a,b}^{HTG} (1)=0,\;\;  \lim_{a,b \rightarrow 0} \log^{HTG}_{a,b} (x) =\ln (x), \\
\log^{HTG}_{a,b}(x) = - \log^{HTG}_{-b,-a}(1/x),\\
\log^{HTG}_{a,0}(x) = \log^{HTG}_{0,a}(x),\\
\displaystyle \frac{1}{\log^{HTG}_{a,b}(1/x)} + \frac{1}{\log^{HTG}_{a,b}(x)} = -(a+b)
\ee

%
In the special case, for $a=1-q$ and $b=0$  (or $a=0$ and $b=1-q$), we obtain the Tsallis $q$-functions
\be
\log^{HTG}_{1-q,0} (x) = \log^{HTG}_{0,1-q} (x) = \log^T_q (x) = \frac{x^{1-q}-1}{1-q} \\
\exp^{HT}_{1-q,0} (x) = \exp^{HTG}_{0,1-q} (x) = \exp^T_q (x) = [1+(1-q)x]_+^{1/(1-q)}
\ee
For $a=-b$ the  HTG $(a,b)$- logarithm and its inverse ($(a,b)$-exponential) can be expressed  via Kaniadakis or Tsallis  deformed functions:
\be
\log^{HTG}_{a,-a} (x) = \frac{1}{a} \; \frac{x^a - x^{-a}}{x^a + x^{-a}}= \frac{1}{a} \; \tanh ( a \; \ln(x)) ,\;\;\ -1/2 \leq a  \leq 1/2, \;\; x>0\\
\exp^{HTG}_{a,-a} (x)= \left [\frac{1+a \; x}{1- a \; x } \right]^{1/(2a)} = \exp^K_a \left(\frac{x}{\sqrt{1-a^2 \; x^2}}\right) =\frac{\exp^T_{1+2a}(x/2)}{\exp^T_{1+2a}(-x/2)}.
\ee

{\bf Remark}: Hanel, Thurner and Gell-Mann introduced the
above HTG Logarithm  in a slightly more general form as \cite{hanel2012}
\be
\displaystyle \log^{HTG}_{a,b,h(x)} (x) = \left( \frac{1}{\frac{2}{a-b} h(\frac{a-b}{2} \ln (x))} -\frac{a+b}{2}\right)^{-1}, \;\; x >0, \;\; a \neq b,
\ee
where $x>0$ and $a,b$ are real hyper-parameters and $h(x): \Real \rightarrow [-1,1]$ is a continuous, monotonically increasing, odd
function with $h'(0)=1$ and $\lim_{x \rightarrow \pm\infty} = \pm1$, for example, $h(x) = \frac{2}{\pi} \arctan (\frac{\pi}{2} x)$ or $h(x) =\tanh(x)$.
Using the de l'H\'opital formula can define the logarithm also for singular values of hyperparameters
\be
 \label{HTG2}
	\log_{a,b,h(x)}^{HTG} (x)=\left\{
	\begin{array}{cl}
\displaystyle \left( \frac{1}{\frac{2}{a-b} h(\frac{a-b}{2} \ln (x))} -\frac{a+b}{2}\right)^{-1}, & x>0, \;\; a \neq b,\\
\\
		\displaystyle \frac{1}{a} h(a \ln(x)) & x>0, \;\; a= -b\\
  \\
	\displaystyle	\frac{\ln(x)}{1-a\ln(x)} &  x>0, \;\; a= b,\\
\\
\displaystyle	\ln(x) &  x>0, \;\; a= b=0
	\end{array}
	\right.
\ee
In the special important  case, by setting $h(x)=\tanh(x)$, we obtain the formula
(\ref{HTG1}).

\subsection{Generalized Multi-Parametric Logarithms}

Let consider the following generating function:
\be
\phi(x) &=&  (\lambda_1 x)^{a} - (\lambda_2 x)^{b} + c x.
\ee
Assuming that $\alpha=1$ and $\sigma \neq 0$, we obtain the following  generalized logarithm
\be
\displaystyle \log_{\sigma,a,b,c,\lambda_1,\lambda_2} (x) = \frac{(\lambda_1 x^{-\sigma})^{a}  - (\lambda_2 x^{-\sigma})^b  - \lambda_1^{a} + \lambda_2^{b} + (c-1) x^{-\sigma} }{(b \lambda_2^{b} - a \lambda_1^{a}) \sigma}, \;\; \text{for}\;\; x>0, \;\; \lambda_1 >0, \lambda_2>0|.
\ee
In the special case,  for $\sigma=-1$, $c=1$ $\lambda_1=\lambda_2=\lambda$ and $a=-b= \kappa$, we obtain the Kaniadakis-Scarfone logarithm:
 \be
\log^{KS}_{\kappa,\lambda} (x) = \frac{2}{\lambda^{\kappa} + \lambda^{-\kappa}} \left[ \log^K_{\kappa} (\lambda x)  - \log^K_{\kappa} (\lambda) \right], \;\; x>0, \;\; \lambda \neq 0, \; \kappa \neq 0.
\label{defgenlogK}
\ee
Taking into account the relationship
\be
\sqrt{1+\kappa^2 \log^2_{\kappa}(\lambda)} = \frac{\lambda^{\kappa} + \lambda^{-\kappa}}{2}
\ee
we have
\be
\log^{KS}_{\kappa,\lambda} (x) = \frac{\log^K_{\kappa} (\lambda x)  - \log^K_{\kappa}(\lambda)}{\sqrt{1+\kappa^2 \log^2_{\kappa} (\lambda)}}.
\ee

In this case, we can easily derive inverse function as
\be
\displaystyle \exp^{KS}_{\kappa,\lambda} (x) = \frac{1}{\lambda} \exp_{\kappa} \left( x \; \sqrt{1+\kappa^2 \log^2_{\kappa}(\lambda)} + \log_{\kappa} (\lambda) \right).
\ee
These functions for $\lambda=1$ simplify to the Kaniadakis $\kappa$-logarithm and $\kappa$-exponential and
for $\lambda \rightarrow 0^+$ or $\lambda \rightarrow \infty$ they  tend to the Tsallis  $q$-logarithm and $q$-exponential.  So the above functions, similar to the KLS $(\kappa,r)$-functions,  interpolate between the Kaniadakis and the Tsallis logarithms and exponentials.

On the other hand, assuming that  $\lambda_1= \lambda_2=1$ and $\sigma=-1, c=1$, we obtain as a special case  the Euler logarithm called also the Borges-Roditi generalized logarithm \cite{Cichocki2025,Borges1998}:
\be
 \log^{E}_{a,b}(x)= \frac{x^a-x^b}{a-b},  \;\; x>0, \;\; a\neq b, \; \; a<1,  \; 0<b<1, \;\; \text{or} \;\;  b< 1, \; 0<a<1,
\label{logab}
 \ee

\section{Properties of  the  Generalized $(\phi, \alpha, \sigma)$-Exponential}

In general, for arbitrary set of parameters, the Tempesta logarithm cannot be inverted analytically, thus it is impossible to define explicitly the corresponding generalized exponential expressed by some basic functions (except of some important special cases discussed above).
However, we  can estimate the value of inverse function by look-up table or approximate it by a power series by applying  Lagrange's inversion theorem (or Lagrange-B\"urmann formula) around $1$ or apply inverse of power series for the Tempesta functional generalized logarithm (\ref{logTS})
Applying the Lagrange-B\"urmann inversion formula, we obtained the following rough power series approximation:

\be
\displaystyle \exp^{Te}_{\phi,\alpha,\sigma} (x) &\approx& 1+x + \frac{1}{2} (1-a_1) x^2 + \left(\frac{1}{2}(1-a_1)^2 + \frac{1}{6} (-2 +3a_1-a_2^2)\right) x^3  +
\cdots  \nonumber\\
&=& 1+x + \frac{1}{2} (1-a-b) x^2 +  \frac{1}{6} \left(1-3a_1-3a^2+3 a_1^2-a_2)\right) x^3  +
\cdots  \nonumber \\
&=& \exp (x) - \frac{1}{2} \left(a_1\right) x^2  -
\frac{1}{6} \left( 3a_1-3a^2+3 a_1^2-a_2 \right) x^3 +  O(x^4),
\ee

The  $(\phi,\alpha,\sigma)$-exponential function should satisfy the following basic properties
\be
\exp^{Te}_{\phi,\alpha,\sigma}(log^{Te}_{\phi,\alpha,\sigma} (x))  &=& x,  \qquad x>0,\\
log^{Te}_{\phi,\alpha,\sigma} (\exp^{Te}_{\phi,\alpha,\sigma}(y))  &=& y, \quad (0 < \exp_{a,b}(y)< +\infty ).
\ee
Furthermore, we have the following fundamental properties.
\be
\frac{d \exp^{Te}_{\phi,\alpha,\sigma}(x)}{dx}&>&0\\
\frac{d^2 \exp^{Te}_{\phi,\alpha,\sigma}(x)}{dx^2}&>&0\\
\exp^{Te}_{\phi,\alpha,\sigma}(0) &=& 1,\\
F(x) &=& \int \log^{Te}_{\phi,\alpha,\sigma} (x) d x  \;\; \text{(It is strictly convex function)}\\
\int_{-\infty}^{0} \exp^{Te}_{\phi,\alpha,\sigma} (x) dx &=& \int_{0}^{1}\log^{Te}_{\phi,\alpha,\sigma} (x) dx > 0,
\ee
The above equations indicate that the generalized  exponential function  is a strictly increasing and convex function, normalized according to $\exp_{\phi,\alpha,\sigma}(0) =1$, and which goes to zero fast enough to be integrable for $x \rightarrow \pm \infty$.

\section{New  Mirror Descent Algorithms Using Multi-parametric Logarithms}


Let assume that the link function is defined as  $f(\bw)= \log_G(\bw) = \log_{\phi,\alpha,\sigma}(\bw)$ and its inverse  $ f^{(-1)} (\bw) =  \exp_G (\bw)= \exp_{\phi,\alpha,\sigma}(\bw)$, then
using a general MD formula (\ref{f-1fDT}), and fundamental properties described above, we obtain a flexible MD  update:
\begin{empheq}[box=\fbox]{align}
\left\{
	\begin{array}{l}
 \displaystyle \tilde{\bw}_{t+1} = \exp^{Te}_{\phi,\alpha,\sigma} \left[\log^{Te}_{\phi,\alpha,\sigma}(\bw_t) - \eta_t \nabla \widehat{L}(\bw_t)\right] \\
\qquad \;\;\;= \bw_t \otimes_{G} \exp^{Te}_{\phi,\alpha,\sigma} \left(-\eta_t \nabla \widehat{L}(\bw_t)\right) \qquad  \text{(Generalized multiplicative update)} \\
 \bw_{t+1} = \frac{\tilde{\bw}_{t+1}}{||\tilde{\bw}_{t+1}||_1}, \;\; \bw_t > 0, \; \forall t
\qquad \qquad \qquad \;\; \text{(Unit simplex projection)}
 \end{array}
	\right.
\end{empheq}
where the generalized  the generalized $G$-multiplication is defined/determined componentwise for two vectors $\bx$ and $\by$ as follows
\be
 \bx \; \otimes_{G} \; \exp^{Te}_{\phi,\alpha,\sigma} (\by) = \exp^{Te}_{\phi,\alpha,\sigma} \left(\log^{Te}_{\phi,\alpha,\sigma} (x) +y\right), \;\; x>0.
\ee
In order to make algorithm stable and to improve its convergence property we use normalized/scaling loss/cost function defined as (see for justification and detail \cite{Cichocki2024})
\be
\widehat{L}(\bw) = L(\bw/||\bw||_1).
\ee

Alternatively, using MMD formula (\ref{diagMD}), we obtain the following additive gradient descent update:
\begin{empheq}[box=\fbox]{align}
\bw_{t+1} &= \left[\bw_t  - \eta_t \diag \left\{\left(
\frac{d\,\log_G(\bw_t)}{d \, \bw_t}\right)^{-1}\right\} \nabla_{\bi w} \widehat{L}(\bw_t)\right]_+, 
\label{diagMDL} \\
\bw_{t+1} &= \frac{\tilde{\bw}_{t+1}}{||\tilde{\bw}_{t+1}||_1}, \quad \quad  \bw_t \in \Real_+^N,\; \forall t \nonumber
\end{empheq}
where $ \displaystyle \diag \left\{ \left(\frac{d\,\log_G(\bw)}{d \bw}\right)^{-1}\right\} = \diag \left\{ \left(\frac{d\,\log_G(\bw)}{d w_1} \right)^{-1}, \ldots, \left(\frac{d\,\log_G(\bw)}{d w_N}\right)^{-1} \right\} $ is a diagonal positive definite matrix.

It is important to note that the above new updates correspond  to a special form of  natural gradient descent w.r.t. the Riemannian metric.  Natural gradient descent is known to be Fisher efficient, meaning it achieves the best possible performance in terms of minimizing the loss function  \cite{Amaribook}. Mirror descent uses the Bregman proximity function to guide the update, while natural gradient descent considers the intrinsic curvature of a Riemannian manifold, using the Fisher Information Matrix to adjust the gradient \cite{MD1}.

\section{Conclusions}

In this paper, we have developed new  generalized and flexible Mirror Descent updates. 
Using generalized entropies and associated deformed multi-parametric logarithms in the Bregman divergence, a regularization term, can provide new insights into mirror descent updates.
The  proposed  family of MD  updates unveil new perspectives on the applications of infinite number of generalized logarithmic and exponential functions and associated generalized entropies, in a wide spectrum of applications formulated as constrained optimization problems. In fact, the proposed MD updates can take a wide variety of different forms depending on selection of hyperparameters and generating functions.
 The hyperparameters associated with deformed logarithm  and exponential functions can be learned, allowing the MD algorithms to adapt to the specific characteristics of the data.
The developed MD algorithms  can be considered as a flexible and robust  alternatives to exponentiated  gradient (EG) descent algorithms for positive weights, especially for sparse data.  The MD algorithms may find potential applications, especially in learning of deep neural networks  and in machine learning  for classification, clustering and predication and in online portfolio selection. Proposed mirror descent algorithms can achieve much faster convergence rates (in some certain cases). They also provide a much more flexible regularization options.

\end{document}